\documentclass{article} 
\usepackage{iclr2017_conference,times}
\usepackage{hyperref}
\usepackage{url}
\usepackage{amsmath,graphicx}
\usepackage{graphicx,epsfig}
\usepackage[utf8]{inputenc} 
\usepackage[T1]{fontenc}    
\usepackage{hyperref}       
\usepackage{url}            
\usepackage{booktabs}       
\usepackage{amsfonts}       
\usepackage{nicefrac}       
\usepackage{microtype}      
\usepackage{epsfig}
\usepackage{alltt}
\usepackage{epstopdf}
\usepackage{multicol}
\usepackage{amsmath}

\usepackage{subfigure}
\usepackage{caption}
\usepackage{bm}

\newcommand{\+}[1]{\bm{#1}} 

\title{Deep Symbolic Representation Learning for \\ Heterogeneous Time-series Classification}

\author{Shengdong Zhang$^{1,2}$, Soheil Bahrampour$^{1}$, Naveen Ramakrishnan$^{1}$, Mohak Shah$^{1,3}$ \thanks{All authors were with the Bosch Research and Technology Center at the time this work is done.} 
\\
$^1$Bosch Research and Technology Center, Palo Alto, CA\\
$^2$Simon Fraser University, Burnaby, BC\\
$^3$University of Illinois at Chicago, Chicago, IL\\
\texttt{sza75@sfu.ca, Soheil.Bahrampour@us.bosch.com}, \\
 \texttt{Naveen.Ramakrishnan@us.bosch.com, Mohak.Shah@us.bosch.com} 
}

\begin{document}

\maketitle

\begin{abstract}
In this paper, we consider the problem of event classification with multi-variate time series data consisting of heterogeneous (continuous and categorical) variables. The complex temporal dependencies between the variables combined with sparsity of the data makes the event classification problem particularly challenging. Most state-of-art approaches address this either by designing hand-engineered features or breaking up the problem over homogeneous variates. In this work, we propose and compare three representation learning algorithms over symbolized sequences which enables classification of heterogeneous time-series data using a deep architecture. The proposed representations are trained jointly along with the rest of the network architecture in an end-to-end fashion that makes the learned features discriminative for the given task. Experiments on three real-world datasets demonstrate the effectiveness of the proposed approaches.
\end{abstract}

\section{Introduction}
\label{sec:intro}
Rapid increase in connectivity of physical sensors and systems to the Internet is enabling large scale collection of time series data and system logs. Such temporal datasets enable applications like predictive maintenance, service optimizations and efficiency improvements for physical assets. At the same time, these datasets also pose interesting research challenges such as complex dependencies and heterogeneous nature of variables, non-uniform sampling of variables, sparsity, etc which further complicates the process of feature extraction for data mining tasks. Moreover, the high dependence of the feature extraction process on domain expertise makes the development of new data mining applications cumbersome. This paper proposes a novel approach for feature discovery specifically for temporal event classification problems such as failure prediction for heating systems.

Feature extraction from time-series data for classification has been long studied~\citep{MM05}. For example, well-known Crest factor~\citep{JN84} and Kurtosis method~\citep{A04} extract statistical measures of the amplitude of time-series sensory data. Other popular algorithms include feature extraction using frequency domain methods, such as power spectral density~\citep{D02}, or time-frequency domain such as wavelet coefficients~\citep{LCLZZ14}.
More recent methods include wavelet synchrony~\citep{MMLK09}, symbolic dynamic filtering~\citep{GR07, BRSDM13} and sparse coding~\citep{HA06, BRSDM13}. On the other hand, summary statistics such as count, occurrence rate, and duration have been used as features for event data~\citep{MHK05}.

These feature extraction algorithms are usually performed as a pre-processing step before training a classifier on the extracted features and thus are not guaranteed to be optimally discriminative for a given learning task. Several recent works have shown that better performance can be achieved
when a feature extraction algorithm is jointly trained along with a classifier in an end-to-end fashion. For example, in~\citet{MBP12, BNRJ16}, dictionaries are trained jointly with classifiers to extract discriminative sparse codes as feature. Recent successes of deep learning methods~\citep{GBC16} on extracting discriminative features from raw data and achieving state-of-the-art performance have boosted the effort for automatic feature discovery in several domains including speech~\citep{KSH12b}, image~\citep{KSH12a}, and text~\citep{SVL14} data. In particular, it has been shown that recurrent neural networks~\citep{E90} and its variants such as LSTMs~\citep{HS97, GMH13} are capable of capturing long-term time-dependency between input features and thus are well suited for feature discovery from time-series data. 

While neural networks have also been used for event classification, these efforts have been mostly focused on either univariate signal~\citep{HCCCD14} or uniformly sampled multi-variate time-series data~\citep{MMLK09}. In this paper, we focus on event classification task (and event prediction task that can be reformulated as event classification), where the application data consists of multi-variate, heterogeneous (categorical and continuous) and non-uniformly sampled time-series data. This includes a wide variety of application domains such as sensory data for internet of things, health care, system logs from data center, etc. Following are the main contributions of the paper:

\begin{itemize}

\item We propose three representation learning algorithms for time-series classification. The proposed algorithms are formulated as embedding layers, which receive symbolized sequences as their input. The embedding layer is then trained jointly with a deep learning architecture (such as convolutional or recurrent network) to automatically extract discriminating representations for the given classification task. The proposed algorithms differ in the way they embed the symbolized data and are named as Word Embedding, Shared Character-wise Embedding, and Independent Character-wise Embedding. 

\item The deep learning architectures combined with the proposed algorithms provide a unified framework to handle heterogeneous time-series data which regularly occur in most sensor data mining applications. They uniformly map data of any type into a continuous space, which enables representation learning within the space. We will provide detailed discussions on the suitability of the proposed representations and their respective strengths and limitations.


\item We show that the proposed algorithms achieve state-of-the-art performance compared to both a standard deep architecture without symbolization and also compared to other classification approaches with hand-engineered features from domain experts. This is shown with experimental results on three real-world applications including hard disk failure prediction, seizure prediction, and heating system fault prediction.
\end{itemize}

\section{Symbolization}
\label{sec:sym}
\vspace{-5pt}

Symbolization has been widely used as first step for feature extraction on time-series data, providing a more compact representation and acting as a filter to remove noise. Moreover, symbolization can be used to deal with heterogeneity of the data where multi-variate time-series contain both categorical, ordinal, and continuous variables. For example, symbolized sequences are used in~\citet{BRSDM13} to construct a probabilistic finite state automata and a measure on the corresponding state transition matrix is then used as final feature which is fed into a classifier. However, this kind of features, which are extracted without explicitly optimizing for the given discriminative task, are typically suboptimal and are not guaranteed to be discriminative. Moreover, incorporating symbol-based representations and jointly training a deep network is non-trivial. In this work, we propose a unified architecture to embed the symbolized sequence as an input representation and jointly train a deep network to learn discriminative features. Symbolization for a discrete variable is trivial as the number of symbols is equal to the number of available categories. For continuous variables, this requires partitioning (also known as quantization) of data given an alphabet size (or the symbol set). The signal space for each continuous variable, approximated by training set, can be partitioned into a finite number of cells that are labeled as symbols using a clustering algorithm such as uniform partitioning, maximum entropy partitioning~\citep{RR06}, or Jenks natural breaks algorithm~\citep{J67}. The alphabet size for continuous variable is a hyper-parameter which can be tuned by observing empirical marginal distributions.

\begin{figure}[!t]
\centering
\includegraphics[height = 3.7cm ,width=2.5in]{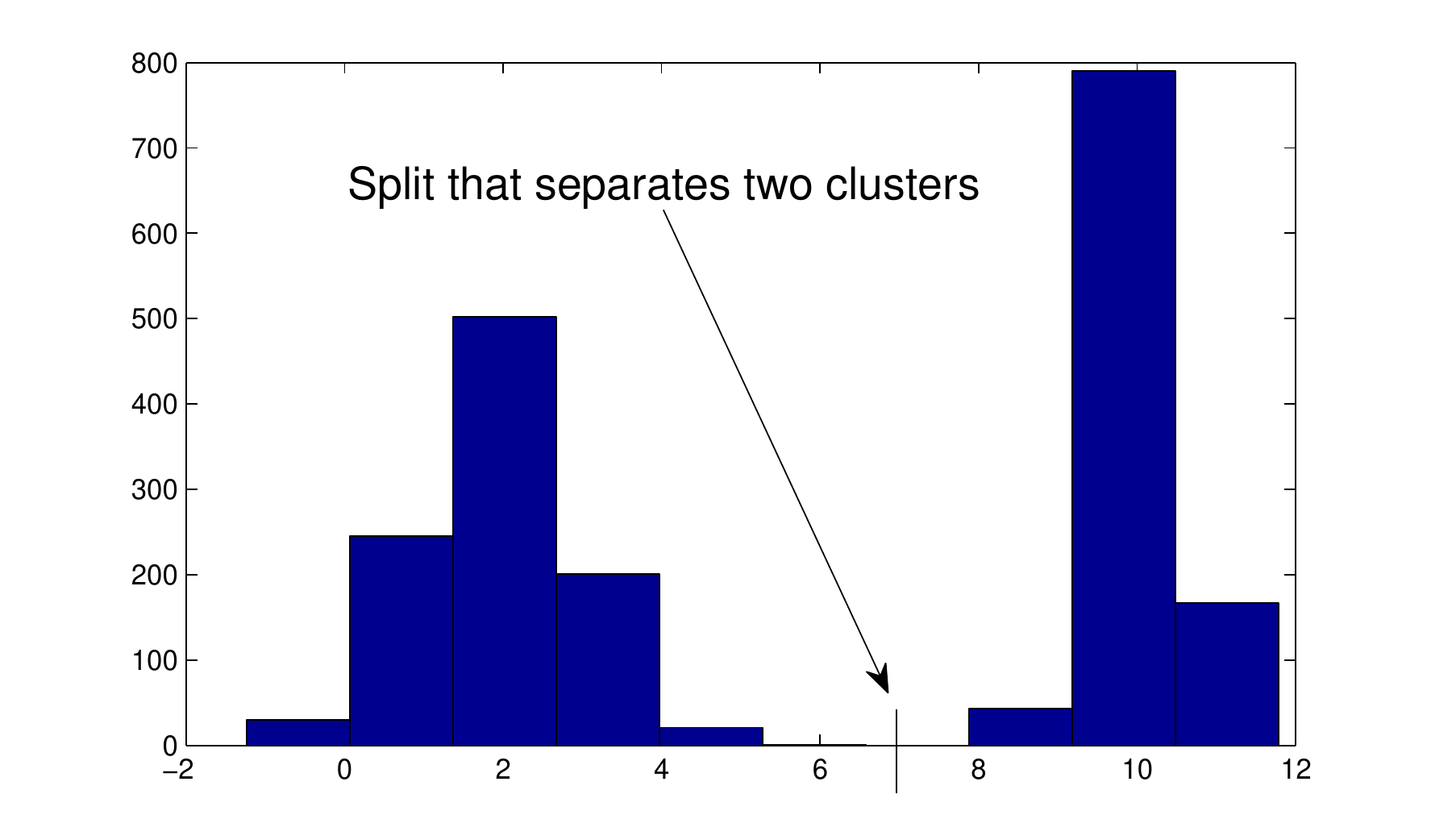}
\caption{Partitiong of continous variable $Z_1$ based on its histogram.}\label{fig:hist} 
\end{figure}

\begin{figure}[!th] 
\begin{center}
\small
$
\begin{array}{ll}
\textrm{time} & \begin{array}{llll}
 \textrm{  }  0\textrm{ } & \textrm{      } 1\textrm{      }  & \textrm{      }2 &  \textrm{      } \cdots \\
\end{array} \\
\begin{matrix}
Z_1\\ 
Z_2\\ 
\end{matrix}
& 
\begin{bmatrix}
2.1 & 11 & 5 & \cdots    \\ 
C_5 & C_2 & C_1 & \cdots \\ 
\end{bmatrix} 
\Rightarrow 
\begin{bmatrix}
a_{Z_1} & b_{Z_1} & a_{Z_1} &\cdots \\ 
e_{Z_2} & b_{Z_2} & a_{Z_2} &\cdots \\ 
\end{bmatrix}
\end{array}
\begin{array}{l}
\nearrow \\ 
\rightarrow \\ 
\searrow 
\end{array}
\begin{array}{l}
\textrm{WdE} ~ \begin{bmatrix}
\+v_{ae} & \+v_{bb} & \+v_{aa} & \cdots
\end{bmatrix}
\\ \\
\textrm{SCE} ~ \begin{bmatrix}
\+v_{a}^1 + \+v_{e}^2& 
\+v_{b}^1 + \+v_{b}^2& 
\+v_{a}^1 + \+v_{a}^2& 
\cdots 
\end{bmatrix} 
\\
\\
\textrm{ICE} ~ \begin{bmatrix}
(v_{a}^{1} \textrm{  } v_{e}^{2})^T& 
(v_{b}^{1} \textrm{  } v_{b}^{2})^T& 
(v_{a}^{1} \textrm{  } v_{a}^{2})^T& 
\cdots
\end{bmatrix}
\end{array}
$
\end{center}
\caption{Heterogeneous time-series symbolization along with word embedding (WdE), shared character-wise embedding (SCE), and independent character-wise embedding (ICE).}\label{fig:symbolization}
\end{figure}

Figures~\ref{fig:hist} and~\ref{fig:symbolization} illustrate the symbolization procedure in a simple example converting a synthetic $2$ dimensional time series $\{Z_1, Z_2\}$ into a sequence of representations. The histogram of continuous variable $Z_1$ contains two Gaussian-like distributions and thus is partitioned into $2$ splits, i.e. for any realization of this variable in the time series, the value is replaced with symbol $a_{Z_1}$ if it is less than $7$, and with $b_{Z_1}$ otherwise. For discrete variable $Z_2$, assuming it has 5 categories $Z_2 \in \{ C_1, C_2, C_3, C_4, C_5\}$, we assign symbol $a_{Z_2}$ to $C_1$, symbol $b_{Z_2}$ to $C_2$, and so on.





\section{Representation Learning}\label{s:RepLearn}

In this section, we propose three methods to learn representation from symbolized data. 

\subsection{Word Embedding (WdE)}\label{ss:word}
Symbolized sequences at each time-step can be used to form a word by orderly collecting the symbol realizations of the variables. Thus, each time-series is represented by a sequence of words where each word represents the state of the multi-variate input at a given time. In Figures~\ref{fig:symbolization}, word embedding vector (WdE) for word $w$ is shown as $\+v_{w}$. Each word of the symbolized sequence is considered as a descriptor of a ``pattern'' at a given time step. Even though the process of generating words ignores dependency among variables, it is reasonable to hypothesize that as long as a dataset is large enough and representative patterns occur frequently, an embedding layer along with a deep architecture should be able to capture the dependencies among the ``patterns''. The set of words on training data construct a vocabulary. Rare words are excluded from the vocabulary and are all represented using a out-of-vocabulary (OOV) word. OOV word is also used to represent words in test set which are not present in training data. 

One natural choice for learning representation of the symbolized sequence is to learn embeddings of the words within the vocabulary. This is done by learning an embedding matrix $\+\Phi \in \mathbb{R}^{d\times v}$ where $d$ is the embedding size and $v$ is the vocabulary size (including OOV word), similar to learning word embedding in a natural language processing task~\citep{DL15}. One difference is that all words here have same length as the number of input variables. Each multi-variate sample is thus represented using a $d$-dimensional vector. It should be noted that the embedding matrix is learned jointly along with the rest of the network to learn discriminative representations. It should also be noted that although the problem of having rare words in training data is somewhat addressed by using OOV embedding vectors, this can limit the representation power if symbolization results in too many low-frequency words. Therefore, the quality of learning with word embeddings highly depends on the cardinality of the symbol set and the splits used for symbolization.


\subsection{Shared Character-wise Embedding (SCE)}\label{ss:c2}

The proposed word-embedding representation can capture the relation among multiple input variables given sufficient amount of training samples. However, as discussed in previous section, the proposed word-embedding representation learning needs careful selection of the alphabet size to avoid having too many low-frequency words and thus is inherently implausible to use in applications where the number of input time-series are too large. In this section, we propose an alternative character-level representation, which we call Shared Character-wise Embedding (SCE), to address this limitation while still being able to capture the dependencies among the inputs.

Instead of forming words at each time step, we use character embedding to represent each symbol and each observation at a time step is represented by the sum of all embedding vectors at a given time step. To formulate this, consider an $m$-dimensional time-series data where the symbol size for the $i$-th input is $s^i$. Let $\+e_l^i \in \mathbb{R}^{s^i}$ be the one-hot representation for symbol $l$ of the $i$-th input and $\+v_{l}^{i}$ be the corresponding embedding vector. Also Let $\Phi = [ \+V^{1} \dots \+V^{m} ] \in \mathbb{R}^{d \times \sum_i{s^i}}$ be the embedding matrix where $\+V^i    \in \mathbb{R}^{d \times s^i}$ is the collection of the embedding vectors for $i$-th input. Then, a given input sample $x^1,x^2,...,x^m$ is represented as $\sum_i{\+V^i\+e^i_{x^i}} \in \mathbb{R}^d$, where $x^i$ is the symbol realization of the $i$-th input. See Figure~\ref{fig:symbolization} for an example of the embedding (SCE) generated using this proposed representation. Since the representation of each word is constructed by summing the embeddings of individual characters, this method does not suffer from the unseen words issues. 

\subsection{Independent Character-wise Embedding (ICE)}\label{ss:c1}
Although SCE does not suffer from the low-frequency issue of WdE, both of these representations do not capture the ordinal information in the input time series. In this section, we propose an Independent Character-wise Embedding (ICE) representation that maintains ordinal information for symbolized continuous variables and categorical variables that have ordered information. To enforce the order constraint, we embed each symbol with a scalar value. Each input $i \in \lbrace 1, \dots, m \rbrace$ is embedded independently and the resulting representation for a given sample $x^1, \dots, x^m$ is $[ \+V^{1}e^1_{x^1} \dots \+V^{m}e^m_{x^m} ]^T \in \mathbb{R}^m$ where $x^i$ is the symbol realization of the $i$th input and $\+V^i$ is a row vector consisting of embedding scalars. The possible correlation among inputs are left to be captured using following layers after the embedding layer in the network. See Figure~\ref{fig:symbolization} for an example of generating embedding vector (ICE) using the proposed algorithm.

\begin{figure*}[!t]
\begin{minipage}{0.4\textwidth}
\centering
\includegraphics[height = 3.3cm ,width=2.3in]{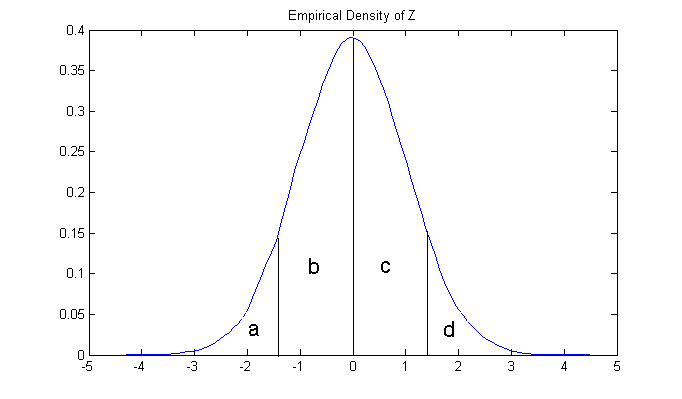}
\end{minipage}%
\begin{minipage}{0.6\textwidth}
\centering
\includegraphics[scale = 0.23]{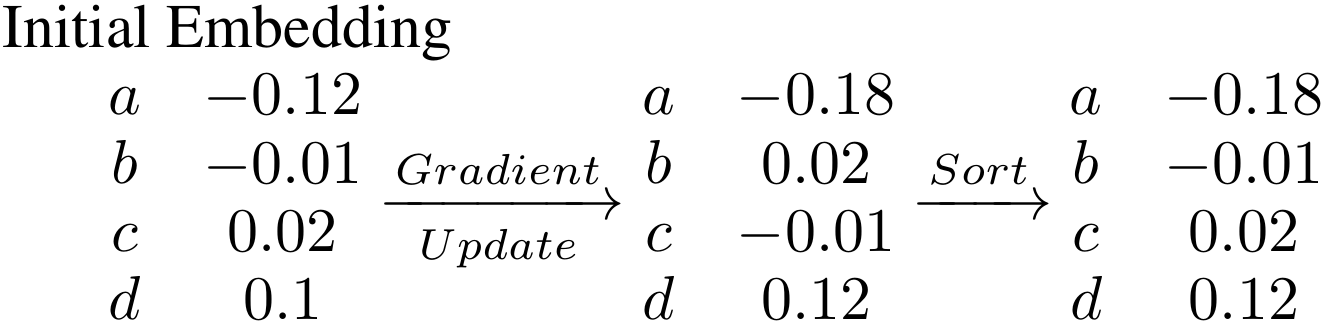}
\end{minipage}
\caption{Empirical Probability Density of an counitous variable $Z$ is shown along with the corresponding independent character-wise embeddings. The vairable has $4$ symbols which are initiazed to maintain the ordered information. During training, and after each gradient update, the representations are sorted to enforce the ordered constraint.}\label{fig:pdf_Z}
\end{figure*}

The embedding scalars for each symbol is initialized to satisfy the ordered information and during training we make sure that the learned representations satisfy the corresponding ordinal information, i.e. the embedding scalars of an ordinal variable are sorted after each gradient update. Figure~\ref{fig:pdf_Z} illustrates this process. 

It should be noted that the embedding layer here has $\sum_is^i$ parameters to learn and thus is slimmer compared to $d\times\sum_is^i$ parameters of the shared character-wise embedding proposed in previous section. Both of the proposed character-wise representations have more compact embedding layer than the word-embedding representation that has $d\times v$ parameters as vocabulary size $v$ is usually large. 



\section{Prediction architecture}
\label{sec:pred_arch}

\vspace{-5pt}

\subsection{Formulation of Prediction Problem}
\label{sec:form_pred}
\vspace{-5pt}

In this section, we formulate the event prediction problem as a classification problem. Let $\+X=\{ \+X_1, \+X_2, \dots, \+X_T \}$ be a time-ordered collections of observation sequences (or clips) collected over $T$ time steps, where $\+X_t=\{\+x_{t1}, \dots, \+x_{tN_t} \}$ represents $t$th sequence consisting of $N_t$ consecutive measurements. As notation indicates, it is \textit{not} assumed that the number of observations within each time step is constant. Let $\{ l_1, l_2, \dots, l_T\}$ be the corresponding sequence labels for $\+X$, where $l_t \in \{0, 1\}$ encodes presence of an event within the $t$th time step, i.e. $l_t =1$ indicates that a fault event is observed within the period of time input sequence $\+X_t$ is collected. We define target labels $y = \{ y_1, y_2, \dots, y_T\}$ where $y_t = 1$ if an event is observed in the next $K$ time-steps, i.e. $\sum_{j=t+1}^{t+K}l_j > 0$, and $y_t = 0$ otherwise. In this formulation, $K$ indicates the prediction horizon and $y_t = 0$ indicates that no event is observed in the next $K$ time-steps, refered to as monitor window in this paper.
The prediction task is then defined as predicting $y_t$ given input $\+X_t$ and its corresponding past measurements $\{\+X_{t-1}, \+X_{t-2}, \dots, \+X_{1} \}$. Using the prediction labels $y$, the event prediction problem on time series data is converted into a classic binary classification problem. Note that although the proposed formulation can in theory utilize \emph{all} the past measurements for classification, we usually fix a window size of $M$ past measurements to limit computational complexity. For instance, suppose that $X_t$'s are sensory data measurements of a physical system collected at the $t$th day of its operation and let $K=7$ and $M=3$. Then the classification problem for $X_t$ is to predict $y_t$, i.e., whether an event is going to be observed in the next coming week of the physical system operation, given current and past three days of measurements. 

\vspace{-5pt}

\subsection{Temporal Weighting Function}
\label{sec: weight_fun}

In rare event prediction tasks, the number of positive data samples, i.e. data corresponding to occurrences of a target event, is much fewer than the one of negatives. If not taken care of, this class imbalance problem causes that the decision boundary of a classifier to be dragged toward the data space where negative samples are distributed, artificially increasing the overall accuracy while resulting in low detection rate. This is a classic problem in binary classification and it is a common practice that larger misclassification cost are associated to positive samples to address this issue~\citep{B01}. However, simply assigning identical larger weights to positive samples for our prediction formulation cannot emphasize the importance of temporal data close to a target event occurrence. We hypothesize that the data collected closer to an event occurrence should be more indicative of the upcoming error than data collected much earlier. Therefore, we design the following weighting function to deal with the temporal importance:

\begin{equation}\label{eq:weight}
w_t = \left\{
\begin{array}{ll}
\sum_{j=1}^{K} (K - j + 1)l_{t+j} & \textrm{if } y_{t}=1 \\
1 & \textrm{if } y_{t}=0
\end{array}
\right.
\end{equation}

This weighting function gives relatively smaller weights to data far from event occurrences compared to those which are closer. In addition to temporal importance emphasis, it also deals with overlapping events. For example, suppose that two errors are observed at time samples $t+1$ and $t+3$ and prediction horizon $K$ is set to 5. Then input sample $X_{t}$ is within the monitor windows of both events and thus its weight is set to higher value of $w_{t} = (5-1+1) + (5-3+1) =8$ as misclassification in this day may result in missing to predict two events. By weighting data samples in this way, a classifier is trained to adjust its decision boundary based on the importance information.

The above weight definition deals with temporal importance information for event prediction. We also need to re-adjust weights to address the discussed class imbalance issue. After determining the weight using Eq.~\ref{eq:weight} for each training sample, we re-normalize all weights such that the total sum of weights of positive samples becomes equal to the total sum of weights of negative samples.

The weighted cross entropy loss function is used as the optimization criterion to find parameters for our model. For the given input $\+X_t$ with weight $w_t$, target label $y_t$, and the predicted label  $\hat{y}_t$, the loss function is defined as :
\begin{equation}
l(y_t, \hat{y}_t) = w_t(y_t log \hat{y}_t + (1-y_t) log (1-\hat{y}_t)).
\end{equation}

\subsection{Network Architecture}
\label{sec:lstm}
Each of the proposed embedding layers can be used as the first layer of a deep architecture. The embedding layer along with the rest of the architecture are learned jointly to optimize a discriminative task, similar to natural language processing tasks~\citep{H16}. Thus, the embedding layer is trained to generate discriminative representations. The specific architectures are further discussed in the results section for each experiment.

\vspace{-10pt}
\section{Results}
\label{sec:results}
\vspace{-5pt}

\subsection{Hard-disk Failure Prediction}
\label{sec:reliability}

Backblaze data center has released its hard drive datasets containing daily snapshot S.M.A.R.T (Self-Monitoring, Analysis and Reporting Technology) statistics for each operational hard drive from 2013 to June 2016. The data of each hard drive are recorded until it fails. In this paper, the 2015 subset of the data on drive model ``ST3000DM001'' are used. As far as we know, no other prediction algorithm has been published on the data set of this model and thus we have generated our own training and test split. The data consists of several models of hard drives. There are $59,340$ hard drives out of which $586$ of them (less than $1\%$) had failed. The data of the following $7$ columns of S.M.A.R.T raw statistics are used: $5, 183, 184, 187, 188, 193, 197$. These columns corresponds to accumulative count values of different monitored features. We also added absolute difference between count values of consecutive days for each input column resulting in overall $14$ columns. The data has missing values which are imputed using linear interpolation. The task is formulated to predict whether there is a failure in the next $K = 3$ days given current and past 4 days data.

 The dataset is randomly split into a training set (containing $486$ positives) and a test set (containing $100$ positives) using hard disk serial number and without loosing the temporal information. Thus training and test set do not share any hard disk. For the experiment using word embedding (WdE), the data are symbolized with splits determined by observation of empirical histogram of every variable. The vocabulary is constructed using all words that have frequency of more than one. The remaining rare words are all mapped to the OOV word resulting into a vocabulary size of 509. For the experiments using shared and independent character-wise embedding, dubbed as SCE and ICE respectively, partitioning is done using maximum entropy partitioning~\citep{BRSDM13} with the alphabet size of 4, i.e. the first split is at the first $25$-th percentile, the second split is at the $50$-th percentile, and so on. The size of the embedding for WdE and SCE are selected as 16 and 2, respectively, using cross validation. Each of the proposed embedding layers is then followed by an LSTM~\citep{HS97} layer with 8 hidden units and a fully connected layer for binary classification. Temporal weighting is not used here as it was not seem to be effective on this dataset, but cost-sensitive formulation is used to deal with this imbalanced dataset. As baseline methods, we also provided the results using logistic regression classification (LR), random forest (RF), and LSTM trained on normalized raw data (without symbolization). For LR and RF, the five days input data are concatenated to form the feature vector. The RF algorithm consists of 1000 decision trees. The LSTM networks are trained using ADAM algorithm~\citep{KB14} with default learning rate of $0.001$. Tabel~\ref{tabel:results_hard_drive} summarizes the performance on test data set. We reported the balanced accuracy, arithmetic mean of the true positive and true negative rates, the area under curve (AUC) of ROC as performance metrics. The balanced accuracy numbers are generated by picking a threshold on ROC that maximizes true prediction while maintaining a false positive rate of maximum $0.05$. As it is seen, the proposed character-level embedding algorithms result in best performances. It should be noted that the input data is summary statistics, and not raw time-series data, and thus as seen the LR and LSTM algorithms, without symbolization, perform reasonably well.

\begin{table}[!t]
\centering
\caption{Performance comparison of fault prediction methods on Backblaze Reliability dataset. Base-line methods include random forest (RF), logistic regression classifier (LR), and LSTM, trained without symbolization. The three proposed embedding representations on symbolized data are WdE-LSTM, SCE-LSTM, and ICE-LSTM.}
\label{tabel:results_hard_drive}
\begin{tabular}{lcc}
Models             & Balanced Accuracy &  AUC of ROC  \\ \toprule
RF        &  0.803  &  0.804 \\
LR        &   0.846  &  0.851\\
LSTM      &  0.832  &   0.865\\
\midrule
WdE-LSTM       & 0.834 &  0.812\\
SCE-LSTM  & {\bf 0.855}  &  0.841 \\
ICE-LSTM &  0.835  & {\bf 0.893} \\
\bottomrule
\end{tabular}
\end{table}

\subsection{Seizure prediction}
\label{ssec:seizure}

We have also compared the performance of the proposed algorithms for seizure prediction. The data set is from the Kaggle American Epilepsy Society Seizure Prediction Challenge 2014 and consists of intracranial EEG (iEEG) clips from 16 channels collected from dogs and human. We used the data collected from dogs in our experiments, not including data from ``Dog$\_$5" as the data from one channel is missing. We generated the test sets from training data by randomly selecting $20\%$ of one-hour clips. The length of each clip is $240,000$. We have down-sampled them to $1,200$ for efficient processing using recurrent networks. Five-fold cross validation is used to report the results. For the WdE algorithm, we observed that $90\%$ of total words generated, using alphabet size of 4, have frequency of one which are all mapped to OOV and has resulted in poor performance. We also observed that using smaller alphabet size for WdE is not helpful resulting in too much loss of information and thus the performance of WdE algorithm is not reported here. For character-level embedding algorithms, maximum entropy partitioning is used with alphabet size of 50. The network used here consists of a one-dimensional convolutional layer with $16$ filters, filter length of $3$ and stride of $1$, followed by a pooling layer of size 2 and stride 2 and an iRNN layer~\citep{LJH15} with $32$ hidden units. We observed better results using iRNN than LSTM and thus we have reported the performance using iRNN. We have also reported the performance using same network on raw EEG data. The performance of these methods are summarized in Tabel~\ref{tabel:results_seizure}. As it is seen, the proposed ICE embedding resulted in the best performance.  

\begin{table}[!t]
\centering
\caption{Performance comparison of the iRNN on raw data as well as iRNN using the proposed character-level embedding methods on symbolized data for seizure prediction.}
\label{tabel:results_seizure}
\begin{tabular}{lcc}
Models             &Balance Accuracy &  AUC of ROC  \\ \toprule
iRNN & 0.669 & 0.69\\
\midrule
SCE-iRNN  & 0.761  & 0.811 \\
ICE-iRNN & {\bf 0.77} & {\bf 0.818} \\
\bottomrule
\end{tabular}
\end{table}

\subsection{Heating System Failure Prediction}
\label{sec:tt_data}
\vspace{-5pt}

\begin{table}[!t]
\centering
\caption{Performance comparison of LR and LSTM on hand-designed features as well as the results generated using LSTM using the proposed embedding methods on symbolized data for fault prediction on Thermo-technology dataset.}
\label{tabel:results_tt}
\begin{tabular}{lcc}
Models             &Balance Accuracy &  AUC of ROC  \\ \toprule
LR  & 0.685 & 0.716 \\
LSTM & {\bf 0.735} & 0.766\\
\midrule
WdE-LSTM  & 0.729  &  0.759 \\
SCE-LSTM & 0.7  &  0.697 \\
ICE-LSTM &  0.733 & {\bf 0.769} \\
\bottomrule
\end{tabular}
\end{table}

We also applied our method on an internal large dataset containing sensor information from thermo-technology heating systems. This dataset contains $132,755$ time-series of $20$ variables where each time-series is data collected within one day. Nine of the variables are continuous and the remaining $11$ variables are categorical. The task is to predict whether a heating system will have a failure in coming week. The dataset is highly imbalanced, where more than $99\%$ of the data have no fault in the next seven days. After symbolizing the training data, for the experiment of using word embedding, the words that have relative frequency of less than $1\%$ are considered as OOV words. The averaged length of each sequence is $6,000$. The embedding dimension for WdE and SCE algorithm are both chosen as 20 using a validation set. The network architecture includes an LSTM layer with 15 hidden units and a fully connected layer of dimension 50 which is followed by the final fully connected layer for binary classification. The same model architecture was used for the experiments of shared and independent character-wise embedding. 

A simple trick is used to increase the use of GPU parallel computing power during the training phase, due to the large size of the training samples. For a given training time-series with $T$ words, instead of sequentially feeding entire samples to the network, time-series is first divided into $M$ sub-sequences of maximum length $\frac{T}{M}$ where each of these sub-sequences are processed independently and in-parallel. A max-pooling layer is then used on these feature vectors to get the final feature vector which represents the entire time-series. The generated feature is fed into a feed-forward neural network with sigmoid activation function to generate the predictions for the binary classification task. We call the sequence division technique as sequence chopping. Even though training an LSTM with this technique sacrifices temporal dependency longer than $\frac{T}{M}$ time steps, we have observed that by selecting a suitable chopping size, we can achieve competitive results and at the same time significant speed-up of the training procedure. The performance of the model having a similar LSTM architecture which is trained on 119 hand-engineered features is reported in Table~\ref{tabel:results_tt}. It should be noted that the hand engineered features have evolved over years of domain expertise and asset monitoring in the field. The results indicates that our methods results in competitive performance without the need for the costly process of hand-designing features. 

\section{conclusions}
\label{sec:conclusions}
\vspace{-5pt}

We proposed three embedding algorithms on symbolized input sequence, namely WdE, SCE, and ICE, for event classification on heterogeneous time-series data.  The proposed methods enable feeding symbolized time-series directly into a deep network and learn a discriminative representation in an end-to-end fashion which is optimized for the given task. The experimental results on three real-world datasets demonstrate the effectiveness of the proposed algorithms, removing the need to perform costly and sub-optimal process of hand-engineering features for time-series classification.



\bibliography{iclr2017_conference}
\bibliographystyle{iclr2017_conference}

\clearpage 

\section{Appendix}\label{sec:appendix}

\begin{table}[!ht]
\centering
\small
\caption{Number of trainable parameters for the three proposed embedding layers as well as total parameters of the network used in the three studied application. For the WdE on heating system data, an approximate number is provided as the vocabulary size is dependent on the training set used among the three cross-validations splits.}
\label{my-label}
\begin{tabular}{lcccccc}
       & \multicolumn{2}{c}{Hard-Disk data} & \multicolumn{2}{c}{Seizure data}  & \multicolumn{2}{c}{Heating System data}                                                                                                     \\ \cmidrule(lr){2-3} \cmidrule(lr){4-5}  \cmidrule(lr){6-7}
Method & \begin{tabular}[c]{@{}c@{}}\#. of embed.\\ parameters\end{tabular} & \begin{tabular}[c]{
@{}c@{}}\#. of total\\ parameters
\end{tabular} &
\begin{tabular}[c]{@{}c@{}}\#. of embed.\\ parameters\end{tabular} & \begin{tabular}[c]{@{}c@{}}\#. of total\\ parameters\end{tabular} & \begin{tabular}[c]{@{}c@{}}\#. of embed.\\ parameters\end{tabular} & \begin{tabular}[c]{@{}c@{}}\#. of total\\ parameters\end{tabular} \\
\midrule
WdE    & 8144                                                               & 8953                                                              & N/A                                                                & N/A                                                               & $\sim$713000                                                       & $\sim$715000                                                      \\
\midrule
SCE    & 112                                                                & 501                                                               & 2400                                                               & 3569                                                              & 1335                                                               & 4046                                                              \\
\midrule
ICE    & 56                                                                 & 815                                                               & 80                                                                 & 2465                                                              & 69                                                                 & 3100                                                              \\
\bottomrule
\end{tabular}
\end{table}

\end{document}